\documentclass{article}

\PassOptionsToPackage{numbers, compress}{natbib}

\usepackage[preprint]{neurips_2019}

\usepackage[utf8]{inputenc} 
\usepackage[T1]{fontenc}    
\usepackage{hyperref}       
\usepackage{url}            
\usepackage{booktabs}       
\usepackage{amsfonts}       
\usepackage{nicefrac}       
\usepackage{microtype}      

\usepackage{tikz} 
\usepackage{pgfplots}

\pgfplotsset{compat=newest} 
\usepgfplotslibrary{units} 

\usepackage{subcaption}

\usepackage{siunitx}
\usepackage{amsmath}
\usepackage{mathtools}

\DeclarePairedDelimiter\floor{\lfloor}{\rfloor}

\title{Mixed Precision Training With 8-bit Floating Point}

\author{%
  Naveen Mellempudi \\
  Parallel Computing Lab, Intel Labs\\
  \texttt{naveen.k.mellempudi@intel.com} \\
\And
  Sudarshan Srinivasan \\
  Parallel Computing Lab, Intel Labs\\
  \texttt{sudarshan.srinivasan@intel.com} \\
\And
  Dipankar Das \\
  Parallel Computing Lab, Intel Labs\\
  \texttt{dipankar.das@intel.com} \\
\And
  Bharat Kaul \\
  Parallel Computing Lab, Intel Labs\\
  \texttt{bharat.kaul@intel.com} \\
}
\begin{document}
\maketitle

\begin{abstract}
Reduced precision computation for deep neural networks is one of the key areas addressing the widening ’compute gap’ driven by an exponential growth in model size. In recent years, deep learning training has largely migrated to 16-bit precision, with significant gains in performance and energy efficiency. However, attempts to train DNNs at 8-bit precision have met with significant challenges because of the higher precision and dynamic range requirements of back-propagation. In this paper, we propose a method to train deep neural networks using 8-bit floating point representation for weights, activations, errors, and gradients. In addition to reducing compute precision, we also reduced the precision requirements for the master copy of weights from 32-bit to 16-bit. We demonstrate state-of-the-art accuracy across multiple data sets (imagenet-1K, WMT16) and a broader set of workloads (Resnet-18/34/50, GNMT, Transformer) than previously reported. We propose an enhanced loss scaling method to augment the reduced subnormal range of 8-bit floating point for improved error propagation. We also examine the impact of quantization noise on generalization and propose a stochastic rounding technique to address gradient noise. As a result of applying all these techniques, we report slightly higher validation accuracy compared to full precision baseline.

\end{abstract}

\section{Introduction}\label{intro}
The unprecedented success of Deep Learning models in a variety of tasks including computer vision\cite{resnet_he}, machine translation\cite{google_gnmt} and speech recognition\cite{speech_hinton},\cite{deep_speech} has led to the proliferation of deeper and more complex models. Algorithmic innovations such as large batch training\cite{keskar2016large} and neural architecture search\cite{neural_arch_search} have enabled models to scale on large compute cluster to accelerate training. This 
enhanced performance has enabled the adoption of larger neural networks. As a consequence, the computational requirements for training Deep Learning models have been growing at an exponential rate\cite{ai_and_compute} over the past few years, outperforming Moore's Law and hardware capabilities by a wide margin.

One of the promising areas of research to address this growing compute gap is to reduce the numeric precision requirements for deep learning. Reduced precision methods exploit the inherent noise resilient properties of deep neural networks to improve compute efficiency, while minimizing the loss of model accuracy. Recent studies\cite{fp16_mixed_nv},\cite{int16_intel} have shown that, deep neural networks can be trained using 16-bits of precision without any noticeable impact on validation accuracy across a wide range of networks. Today, state-of-the-art training platforms support 16-bit precision in the form of high-performance systolic array or GEMM engine (General Matrix Multiply) implementations\cite{nvidia_tensorcores}, \cite{flexpoint}.

There have been numerous attempts \cite{hubara_bnn},\cite{zhou2016dorefa},\cite{8b_halp},\cite{iclr17_int8},\cite{halfwave_gaussian} to train deep neural networks at lower precision (< 16-bits) with varying degrees of success. With the abundance of 8-bit integer deep learning `ops' deployed to accelerate inference tasks, much of the research into training methods have also focused on integer based fixed-point numeric formats\cite{zhou2016dorefa},\cite{8b_halp},\cite{iclr17_int8}. Training with 8-bit integers has been significantly more challenging because the dynamic range of such formats is not sufficient to represent error gradients during back-propagation. More recently, Wang et al.\cite{NIPS2018_fp8} have shown that 8-bit floating representation can be used to train convolutional neural networks, with the help of specialized chunk-based accumulation and stochastic rounding hardware. While this method has shown promising results, it requires expensive stochastic rounding hardware built into the critical compute path making it unattractive for systolic array and GEMM accelerator implementations.    

Our paper extends the state of the art in 8-bit floating point (FP8) training with the following key contributions:

\begin{itemize}
\item We propose a simple and scalable solution for building FP8 compute primitives, eliminating the need for stochastic rounding hardware in the critical compute path, as proposed in \cite{NIPS2018_fp8}, thereby reducing the cost and complexity of the MAC unit.    

\item Demonstrate state-of-the-art accuracy using 8-bit floating point representation for weights, activations, errors and weight gradients, across multiple data sets (Imagenet-1K, WMT16) and a broader set of workloads (Resnet-18/34/50\cite{resnet_he}, GNMT\cite{google_gnmt}, Transformer\cite{attention_transformer}) than previously reported\cite{NIPS2018_fp8}. We also reduce the precision requirements for the master copy of weights from 32-bit to 16-bit reducing memory footprint of the model by half.  

\item Propose enhanced loss scaling method to compensate for the reduced subnormal range of 8-bit floating point representation for improved error propagation leading to better model accuracy. 

\item Present a detailed study of the impact of quantization noise on model generalization and propose a stochastic rounding technique to address the gradient noise in the early epochs leading to better generalization. As a result of this technique, we even report slightly higher validation accuracy compared to our full precision baseline.

\end{itemize}

\section{Related Work}\label{rel_work}
The study of reduced precision methods for deep learning training is an active area of research. In the pursuit of improving compute efficiency, researchers have experimented with various numeric formats and hardware implementations. Gupta et al.\cite{gupta_16bit} demonstrated that deep neural networks can be trained with minimal loss in accuracy, using 16-bit fixed point representation. This was followed by studies employing other numeric formats such as, half-precision floating point\cite{fp16_mixed_nv} and dynamic fixed point \cite{koster2017flexpoint}, \cite{int16_intel}, demonstrating state of the art results across residual\cite{resnet_he}, recurrent\cite{google_gnmt} and generative networks. Today most of the neural network training in a production deployment has migrated to 16-bit hardware, resulting in significant improvements\cite{nvidia_tensorcores} in performance. 

There have been several attempts to further reduce the precision requirements of DNNs to boost training performance. DoReFa-Net\cite{zhou2016dorefa}, a derivative of AlexNet\cite{alexnet} was trained using bit-convolutions with 1-bit and 2-bits to represent weights and activations respectively, while the gradients are quantized to 6-bits of precision. Wu et al.\cite{iclr17_int8} have trained AlexNet\cite{alexnet} using 8-bit precision for activations, errors and weight gradients, while the weights are quantized to 2-bits of precision. However, both these methods have reported significant loss in validation accuracy.

More recently, Wang et al.\cite{NIPS2018_fp8} have successfully trained Resnet-50\cite{resnet_he} using 8-bit floating point  numeric format with the help of a specialized hardware to compute chunk-based dot-product computation and stochastic rounding on a 16-bit accumulator. The authors of this study have focused on reducing the accumulator precision and based on studies on smaller networks (AlexNet Resnet-18), attributed training issues related to error propagation 
and generalization on the choice of accumulator size. 
However, our studies on larger networks (Resnet-34/50) using 32-bit accumulator for dot-product computations indicate that, these issues are not related to the choice of accumulator size and should be addressed independently. 
We discuss these issues and our proposed solutions in greater detail in Sections\ref{loss_scaling}and \ref{noise_general}. Guided by these results, we decided to focus on studying the impact of using FP8 numeric format on training, while maintaining a high precision accumulator(FP32). We further believe that modern GEMM engine designs implementing progressive multiplier reduction\cite{pmr} techniques can effectively amortize the cost of a larger final accumulator, and do not benefit significantly from 16-bit solutions\cite{NIPS2018_fp8} with additional overheads of chunk-based accumulation and stochastic rounding in the critical path. 
\section{Training Method}\label{method}

The choice of bit-level representation of floating point \textbf{(sign, exponent, mantissa)}, has a significant impact on the effectiveness of the numerical format -- the trade-off between the dynamic range and precision is especially tricky at low bit-width representations. While it is important to maintain higher dynamic range for effective propagation of error gradients\cite{fp16_mixed_nv}, it leads to having values that are too few and scattered to maintain fidelity required for gradient computations. 
After careful consideration of these facts and several failed experiments with other formats (for example with more exponent bits), we decided to use \textbf{s=1,e=5,m=2} numeric format for representing 8-bit floating point.
We also decided to use a 32-bit floating point accumulator; therefore each tensor GEMM/convolution operation takes two input tensors in 8-bit floating point format and produces a 32-bit single precision floating point output.
At this stage the 32-bit output must be down-converted to a 8-bit value in order to be used by the next operation. 
Here, we believe rounding plays an extremely important role and helps recover the numeric accuracy of key compute primitives used by deep learning applications.  
We present the results from the study of different rounding modes applied to this format and their impact on training in Section.\ref{noise_general}

\begin{figure}[htp]
\centering
\begin{subfigure}{.495\textwidth}
  \centering
  \setlength{\fboxsep}{2pt}\fbox{\includegraphics[width=0.97\linewidth,keepaspectratio]{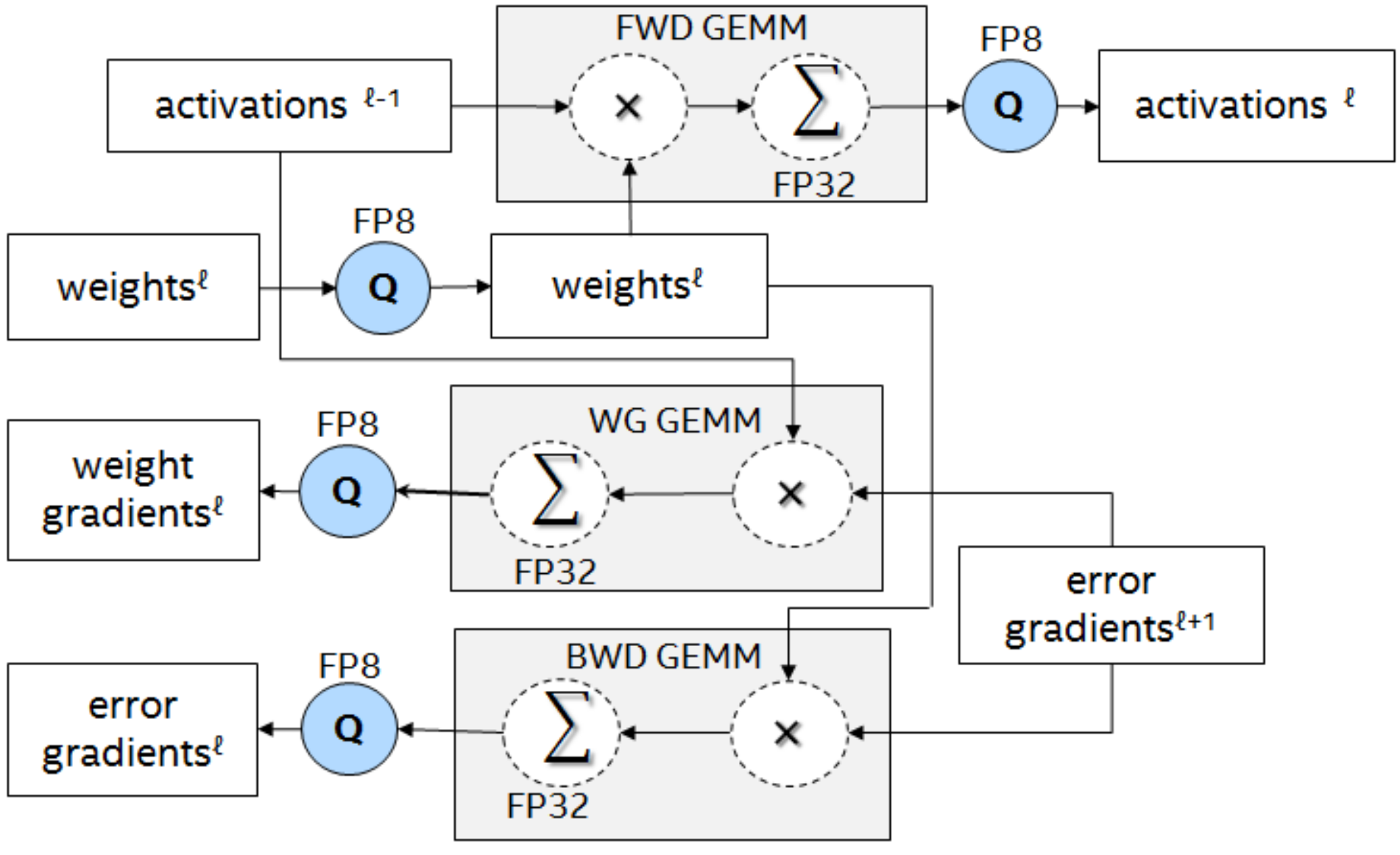}}
 \caption{}
  \label{fig:dataflow_a}
\end{subfigure}
\begin{subfigure}{.495\textwidth}
  \centering
  \setlength{\fboxsep}{2pt}\fbox{\includegraphics[width=0.97\linewidth,keepaspectratio]{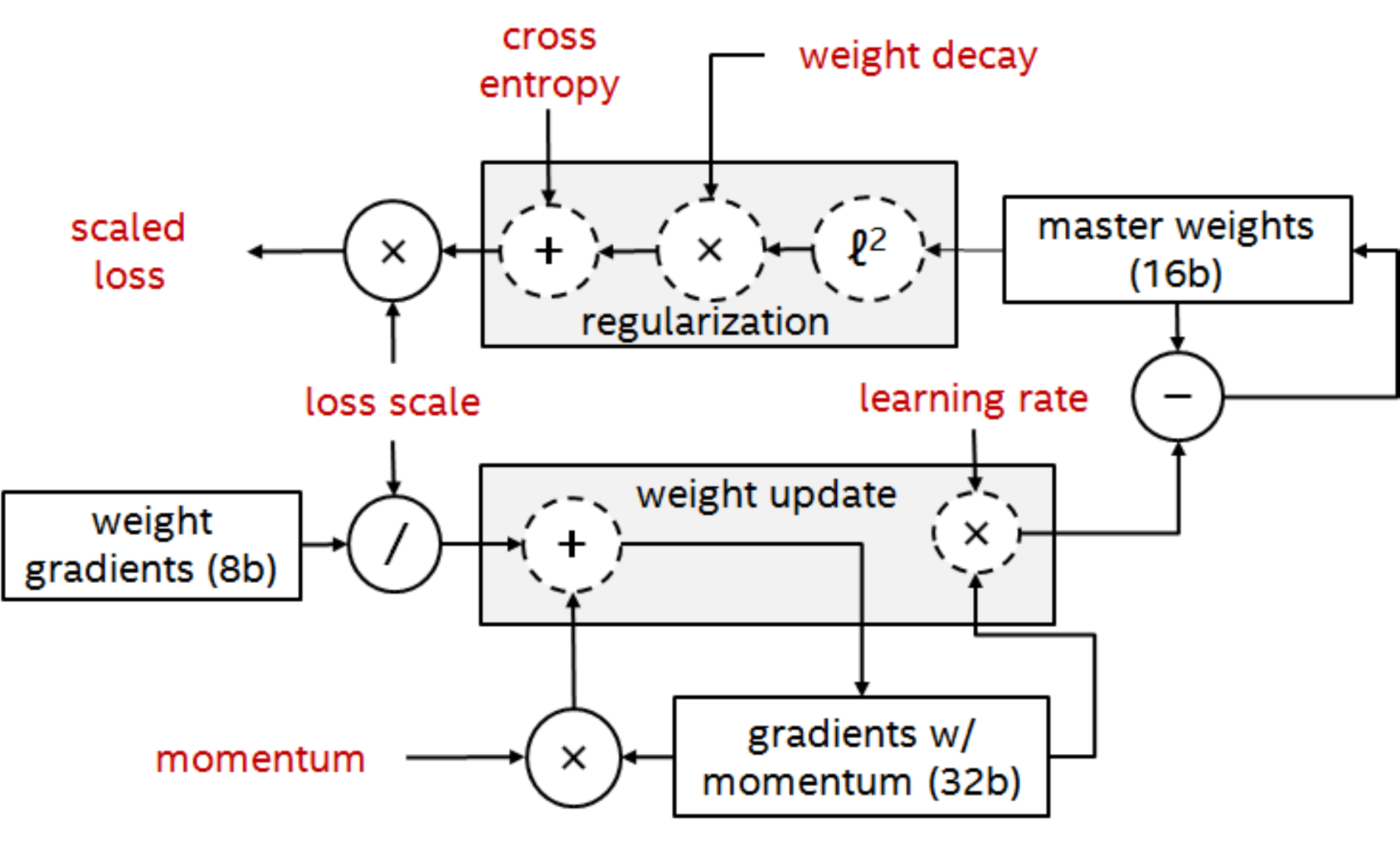}}
  \caption{}
  \label{fig:dataflow_b}
\end{subfigure}
\caption{Mixed precision data flow for FP8 training. (left) precision settings for key compute kernels in Forward, Backward, and Weight Update passes, (right)  flow diagram of the weight update rule.}
\label{fig:data_flow}
\end{figure}

Figure.\ref{fig:data_flow} shows the precision settings of various compute operations used in our mixed precision training setup. The \textbf{'GEMM'}(matrix multiply) operator shown in Figure.\ref{fig:dataflow_a} represents the key compute kernel used by deep neural networks during forward, backward, and gradient computation passes. Quantization nodes identified with the letter \textbf{'Q'} perform down-conversion and rounding operations on the 32-bit floating point output to convert to 8-bit format before passing on to the next layer. For our experiments, we convert the weights, activations, error and weight gradients of all convolution and GEMM kernels to 8-bit floating point format for forward, backward and weight update paths. 

Figure.\ref{fig:dataflow_b} shows the data flow during optimization and weight update steps. In the optimization path the L2-regularization term is added to the cross entropy. Then the loss value is scaled with loss scaling factor before initiating back propagation. When the back propagation is complete the weight gradients are computed and stored in 8-bit floating point format. To perform weight update operation, first the 8-bit weight gradients need to be scaled back by dividing the weight gradients with 'loss scale' parameter. This step is performed in full precision to prevent underflow. The gradients are then passed to the momentum optimizer, the final gradients are then applied to the master copy of the weights. For our experiments, we use half-precision floating point format to store master copy of weights. During the update step, these half precision values are up-converted to 32-bit while they are loaded into the compute unit. The weight update operation is performed as a 32-bit operation. After the update, the master weights are converted back to 16-bit format before they are stored back into memory. Since this is a bandwidth bound operation, performing the update operation in FP32 will not have any noticeable impact on the performance.


\subsection{Enhanced Loss Scaling}\label{loss_scaling}
Previous studies\cite{fp16_mixed_nv} on half-precision floating point have shown that loss scaling technique can be used to push smaller error gradients into representable range and train neural networks successfully. The full range of numeric values represented by a floating point format include the 'subnormal' values, the range of which is determined by the number of mantissa bits. Because of this property of floating point numbers, the proposed 8-bit floating point format will have significantly smaller subnormal range compared to a half-precision floating point with the same number of exponent bits. Table.\ref{tab:float_range} shows the dynamic range comparison between full-precision(FP32), half-precision(FP16) and the proposed 8-bit floating point formats.

\begin{table}[h] 
  \caption{Dynamic range comparison between proposed FP8 and other existing floating point formats.}
  \label{tab:float_range}
  \centering
  \begin{tabular}{lllll}
    \toprule
    Data Type   & Bit Format (s, e, m)    & Max Normal   & Min Normal     & Min Subnormal   \\
    \midrule
    \centering
    IEEE-754 float     & 1, 8, 23  & \num{3.40e+38} & \num{1.17e-38}  & \num{1.40e-45} \\
    IEEE-754 half-float     & 1, 5, 10 & \num{65535}  & \num{6.10e-5}  & \num{5.96e-8}  \\
    FP8 (proposed)      & 1, 5, 2   & \num{57344}  & \num{6.10e-5}  & \num{1.52e-5}   \\
    \bottomrule
  \end{tabular}
\end{table}

\begin{figure}[htp]
\centering
\begin{subfigure}{.495\textwidth}
  \centering
  \setlength{\fboxsep}{2pt}\fbox{\includegraphics[width=0.97\linewidth,keepaspectratio]{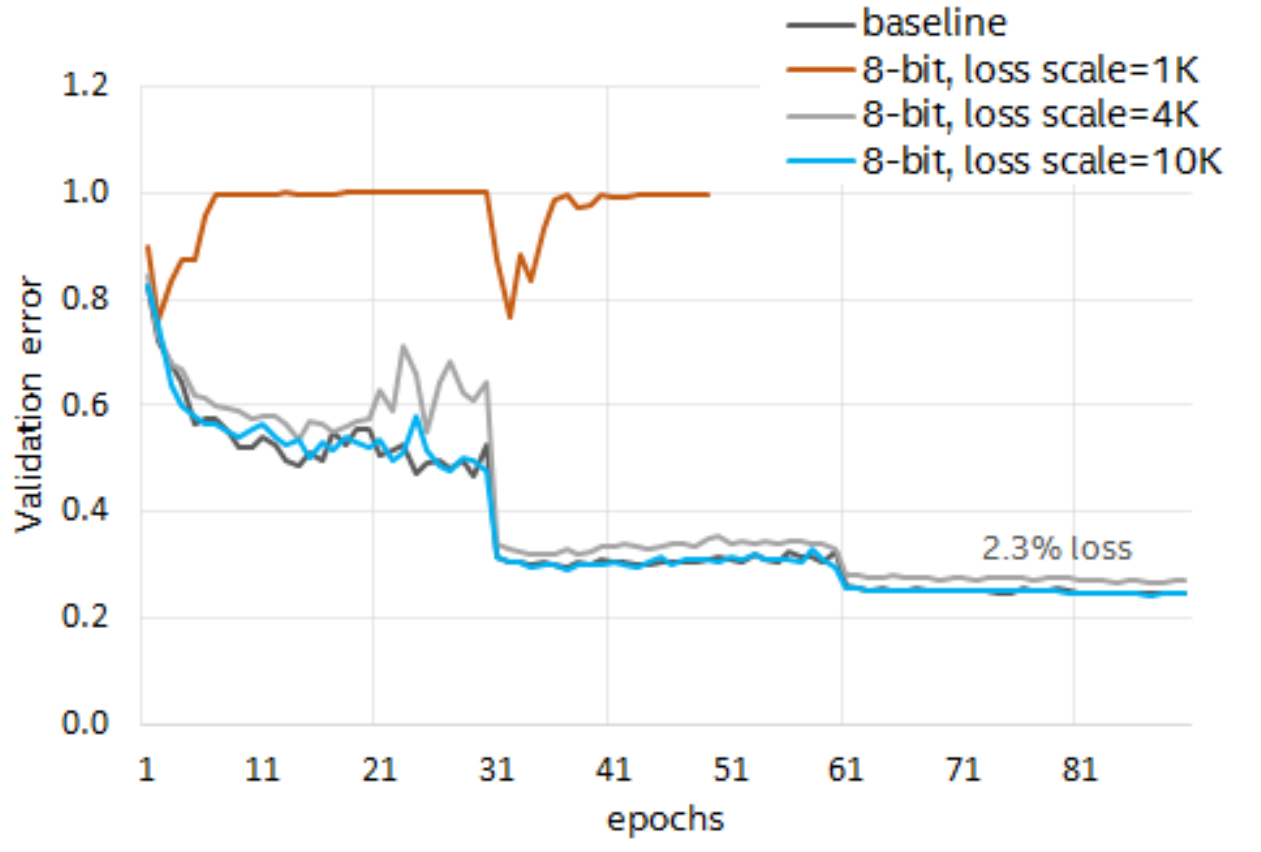}}
 \caption{}
  \label{fig:loss_scale_resnet}
\end{subfigure}
\begin{subfigure}{.495\textwidth}
  \centering
  \setlength{\fboxsep}{2pt}\fbox{\includegraphics[width=0.97\linewidth,keepaspectratio]{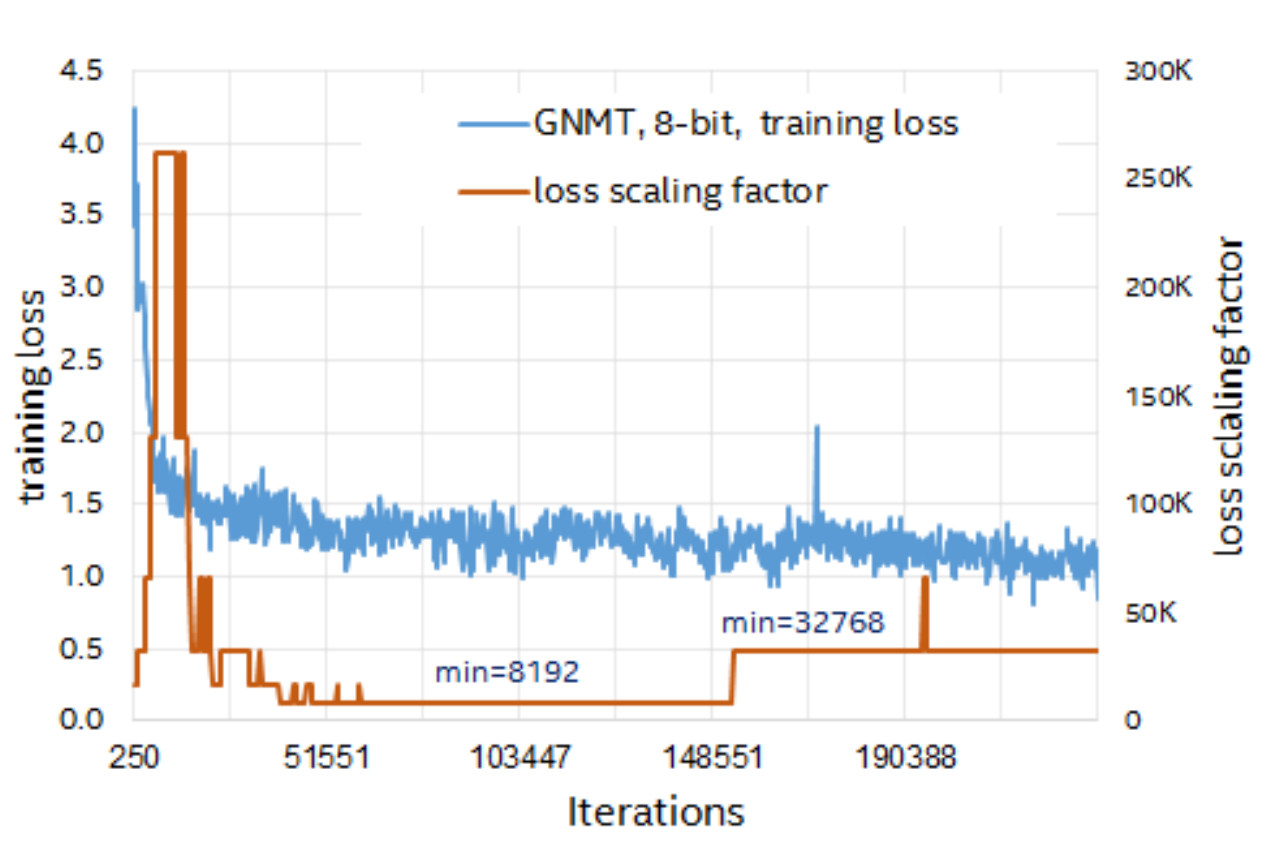}}
  \caption{}
  \label{fig:loss_scale_gnmt}
\end{subfigure}
\caption{Convergence behaviour of FP8 training using enhanced loss scaling. (left) Resnet-50\cite{resnet_he} failed to converge with loss scale=\num{1000}, performed better with \num{2.3}\% accuracy loss at loss scale=\num{4000} and showed full convergence at loss scale=\num{10000}, (right) Dynamic loss scaling with gradually increasing minimum threshold for the scaling factor.}
\label{fig:loss_scaling}
\end{figure}

Half-precision training for convolution networks has been shown to converge using a constant loss scaling parameter of \num{1000}\cite{fp16_mixed_nv}. Other networks such as GNMT\cite{google_gnmt} and Transformer\cite{attention_transformer} use a more robust dynamic loss scaling method\cite{kuchaiev2018openseq2seq}. However, the reduced subnormal range of 8-bit floating point presents a few additional challenges to these methods. For convolution networks, simply increasing the scaling factor addresses the issue of convergence. Figure.\ref{fig:loss_scale_resnet} shows results from our convergence studies on Resnet-50 using different loss scaling values. The model failed to converge with a scaling factor of \num{1000}, and progressively performed better with increasing loss scale values, converging at \num{10000}. Recurrent networks like GNMT\cite{google_gnmt} experience significant variations in gradient distributions through the training cycle and are more sensitive to numerical errors. We trained GNMT using 'back-off' dynamic loss scaling method\cite{kuchaiev2018openseq2seq} which updates the scaling factor every few iterations. While this method is effective in preventing overflows, it proved less effective in handling underflow that occurs more frequently during FP8 training. Our experiments with more frequent updates to scaling factor led to unstable loss behaviour resulting in divergence. We addressed this by gradually increasing the 'minimum threshold' value of the scaling factor by observing the loss function as the training progressed. Figure.\ref{fig:loss_scale_gnmt} shows the loss scaling schedule that worked for GNMT -- we set the minimum threshold to 8K after the first 40K iterations, then increased it to 32K at around 150K iterations. 

\subsection{Quantization noise and Generalization}\label{noise_general}
Reduced precision methods introduce significant amount of noise that can adversely effect convergence and accuracy of deep neural networks. Rounding techniques applied to quantization methods can be effective in regulating some of this noise. For extremely low precision representations with large rounding errors such as the one proposed here($\epsilon = 0.125$), the choice of rounding method can have significant influence on the numeric accuracy and overall applicability of the numeric format. Previous studies\cite{gupta_16bit} have shown that stochastic rounding can be effective for training neural networks using low-precision fixed point formats. 
The most widely supported rounding method in hardware today is RNE (round to nearest even), because it is easier to implement and requires smaller silicon area. In this section, we explore the impact of both RNE and stochastic rounding methods on model convergence and generalization. 

\begin{figure}[htp]
\centering
\begin{subfigure}{.327\textwidth}
  \centering
  \setlength{\fboxsep}{0pt}\fbox{\includegraphics[width=\linewidth,keepaspectratio]{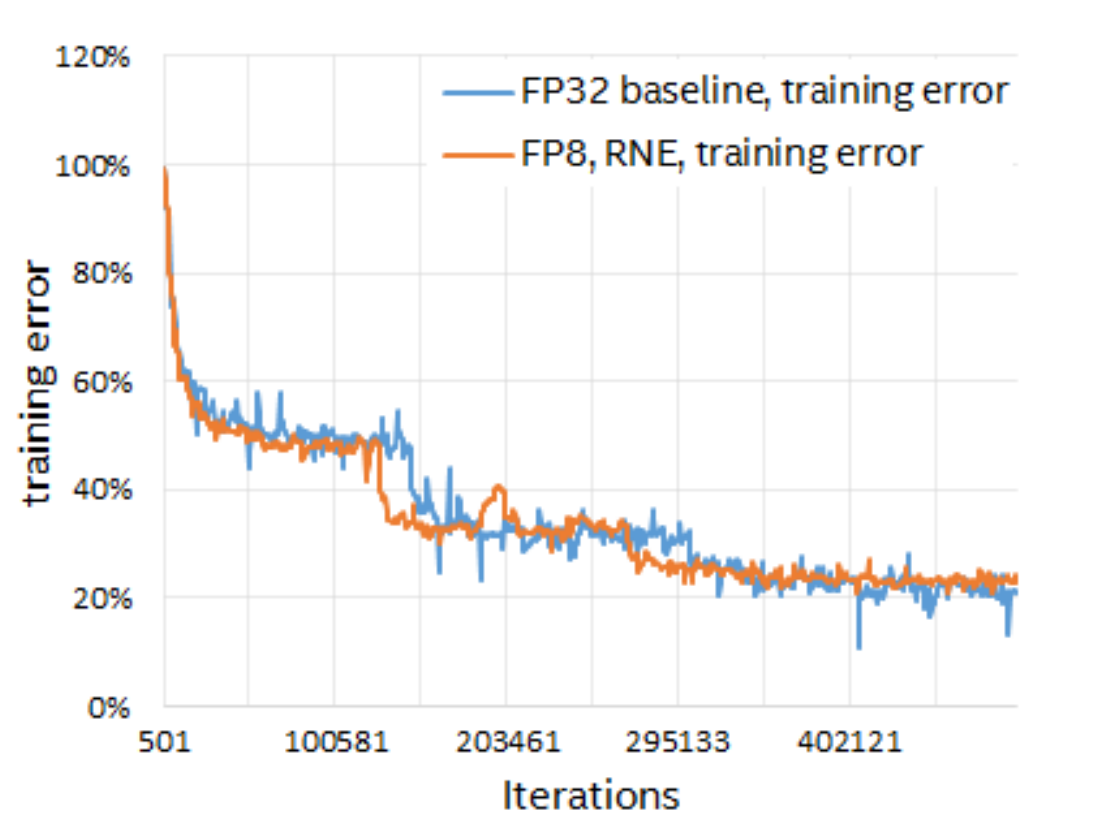}}
 \caption{}
  \label{fig:gen_gap_1}
\end{subfigure}
\begin{subfigure}{.327\textwidth}
  \centering
  \setlength{\fboxsep}{0pt}\fbox{\includegraphics[width=\linewidth,keepaspectratio]{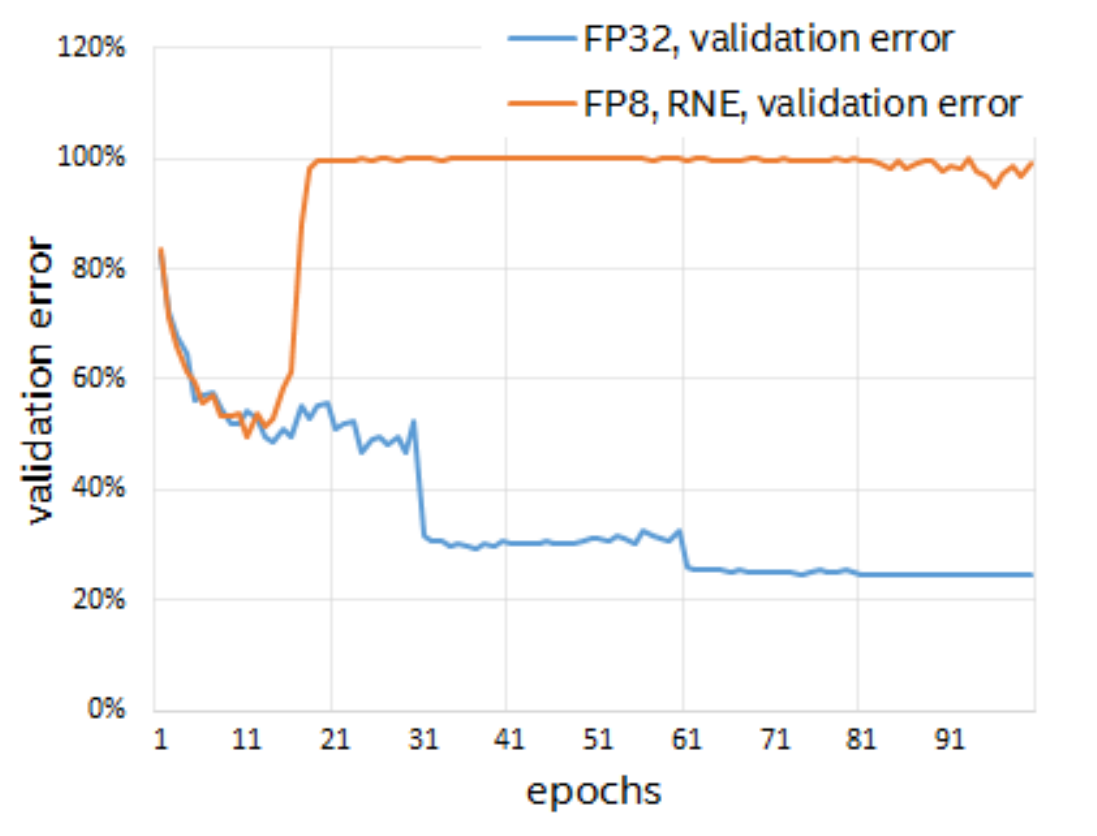}}
  \caption{}
  \label{fig:gen_gap_2}
\end{subfigure}
\begin{subfigure}{.327\textwidth}
  \centering
  \setlength{\fboxsep}{0pt}\fbox{\includegraphics[width=\linewidth,keepaspectratio]{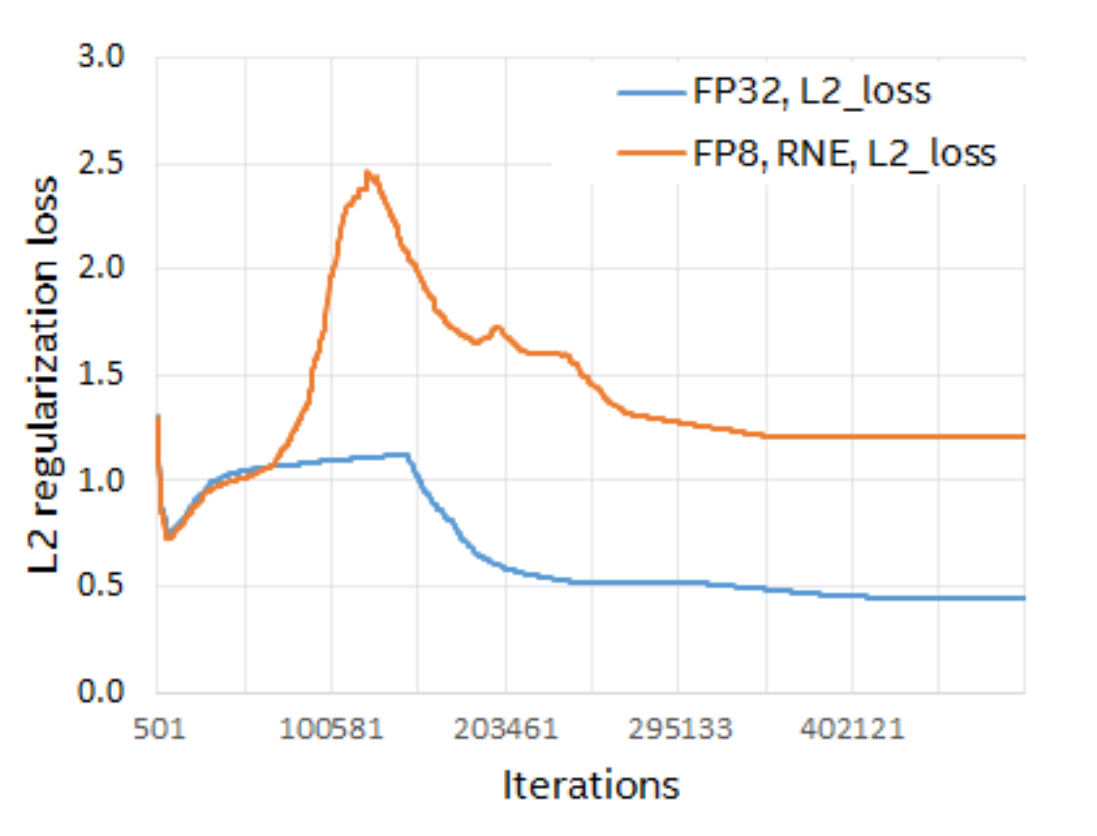}}
  \caption{}
  \label{fig:gen_gap_3}
\end{subfigure}
\caption{Impact of quantization (with RNE rounding) noise on model convergence with Resnet-50 (a) comparison of training error,  (b)  validation error, and (c) L2 regularization loss with FP32 baseline.}
\label{fig:gen_gap}
\end{figure}

Our early experiments showed that, for smaller networks such as Resnet-18\cite{resnet_he}, RNE proved quite effective when trained on Imagenet-1K\cite{imagenet_cvpr09} data set. However, when we trained ResNet-50\cite{resnet_he} we observed some interesting results. Figure.\ref{fig:gen_gap} shows the convergence plots for Resnet-50\cite{resnet_he} using RNE rounding method applied to quantized weights, activations and gradients. The model displayed significant over-fitting behaviour as indicated by the increased validation error, while the training error mostly follows the baseline as shown in as shown in Figure.\ref{fig:gen_gap_2}, and \ref{fig:gen_gap_1}. Multiple experiments indicated that this  behaviour is caused by the noisy error gradients during early epochs which lead to unconstrained growth in model parameters. This is indicated by steep increase in L2 regularization parameter as shown in Figure.\ref{fig:gen_gap_3}. Regularization loss is computed using the formula shown in Equation.\ref{eq:l2_loss}. Increased regularization loss leads to more noisy gradients, which further exacerbates this behaviour. 

An interesting observation about the L2 regularization loss is that for ResNet-18,  the L2-loss is low at the beginning and increases with gradually with iterations. On the other hand for ResNet-50, the L2-loss is high at the beginning due to the initialization of low fan-in 1x1 \cite{Glorot} convolutions, and needs to dip a little before gradually rising again. We suspect that this property of the initialization leads to more noisy behavior of ResNet-50 in the earlier iterations as compared to ResNet-18. Therefore for the ResNet-50 model stochastic rounding is essential.

\begin{equation}\label{eq:l2_loss}
 L2\textunderscore loss = \lambda\times\sum_{i=0}^{W} w_i^2
\end{equation}
Where, $\lambda$ is the weight decay parameter and W is the total number of weights.

\begin{figure}[htp]
\centering
\begin{subfigure}{.495\textwidth}
  \centering
  \setlength{\fboxsep}{2pt}\fbox{\includegraphics[width=0.97\linewidth,keepaspectratio]{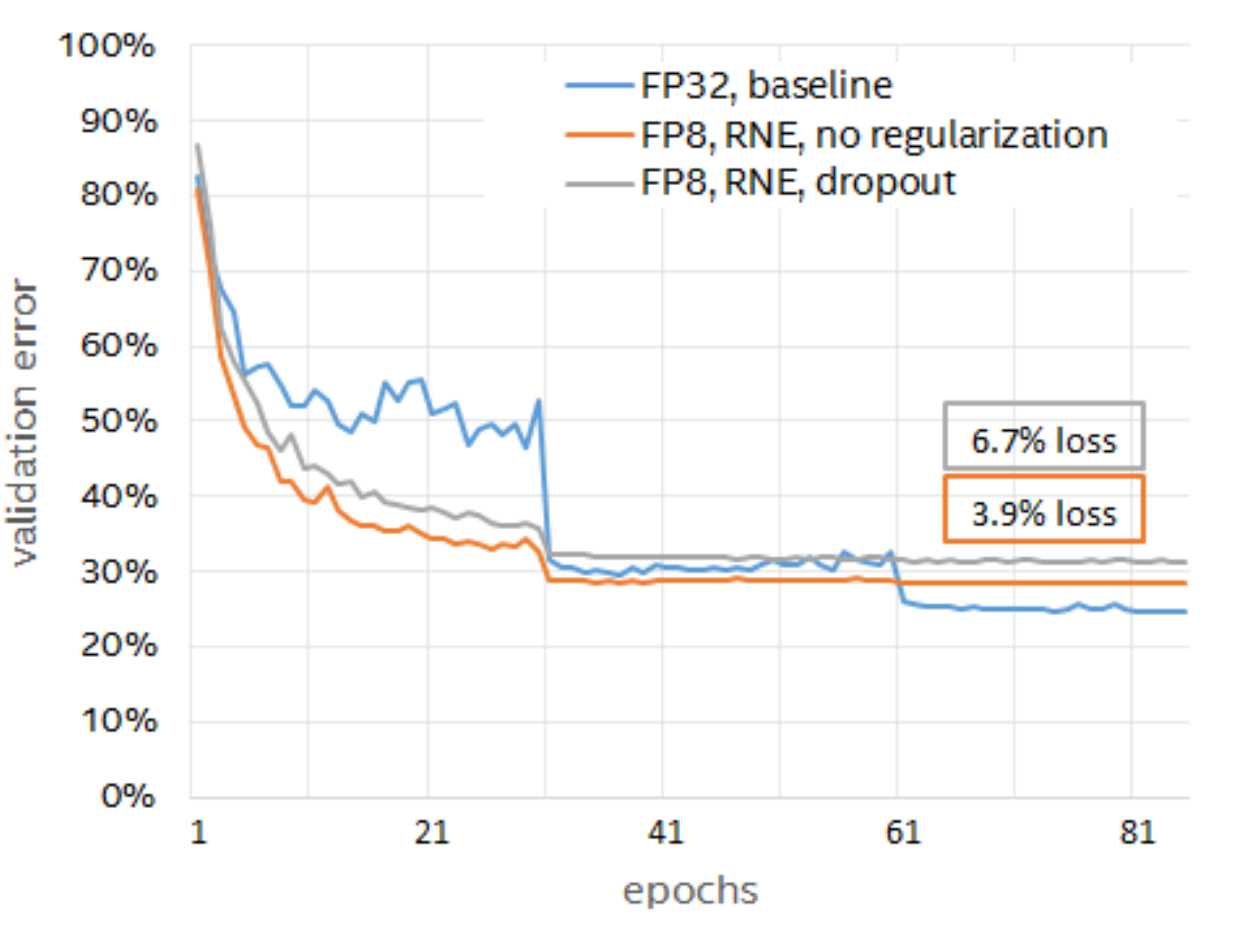}}
 \caption{}
  \label{fig:dropout_plot}
\end{subfigure}
\begin{subfigure}{.495\textwidth}
  \centering
  \setlength{\fboxsep}{2pt}\fbox{\includegraphics[width=0.97\linewidth,keepaspectratio]{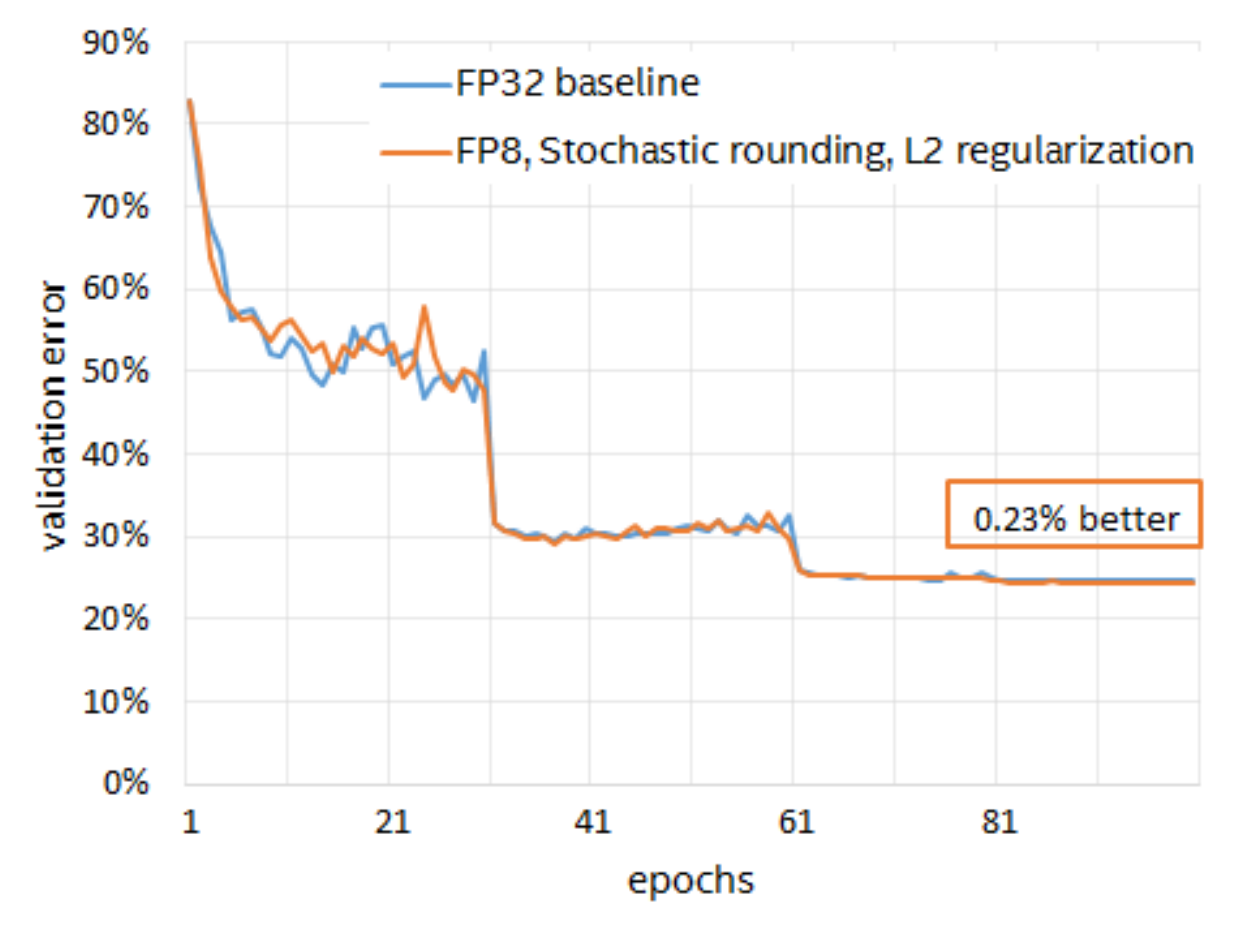}}
  \caption{}
  \label{fig:stochastic_plot}
\end{subfigure}
\caption{(a) Comparing validation performance with 'dropout' and noise-based implicit regularization techniques using RNE(round to nearest even) (b) model  performance with stochastic rounding with L2 regularization.}
\label{fig:alternate_regularization}
\end{figure}

In order to understand the issue of regularization independent of the choice of rounding method, we conducted additional experiments using RNE with other forms of regularization. Figure.\ref{fig:dropout_plot} compares the 'Dropout' method with 'no regularization' method which uses quantization noise as implicit regularizer with no explicit regularization term. In both these cases, the models performed much better than using L2 regularization with RNE, leading us to the conclusion that RNE was ineffective in regulating quantization noise in gradients causing unconstrained growth in model parameters.

Unlike deterministic rounding techniques, stochastic rounding computes the probability of rounding using information from several discarded bits of the input making it less prone to introducing large rounding errors. We studied the error behaviour of Resnet-50\cite{resnet_he} by applying stochastic rounding on activations and gradients to regulate quantization noise in the gradients, which in-turn can improve the effectiveness of explicit regularization methods. 

Our stochastic rounding method is defined as follows:  
\[
    round(x, k)= 
\begin{cases}
    \floor*{x}_k + \epsilon, & \text{with probability } P=\frac{(x-\floor*{x}_k) + r}{\epsilon}\\
    \floor*{x}_k,           & \text{with probability } 1-P
\end{cases}\\
\]
Where, k is the target precision, $\epsilon$ is machine epsilon, and $r$ is random value generated by a pseudo random number generator.

Figure.\ref{fig:stochastic_plot} shows the results from Resnet-50\cite{resnet_he} training experiment using a combination stochastic rounding and explicit L2 regularization. The convergence plots show the good generalization behavior that tracks with the full precision training. As a positive side effect, we have also observed that this method consistently outperformed leading to slightly better validation accuracy across convolution networks. The accuracy numbers are summarized in Section.\ref{experiments}.       

\section{Experiments and Results}\label{experiments}
We built a TensorFlow based training platform\cite{tf_framework}, that can accurately emulate the numeric properties of 8-bit floating point on the current generation floating point hardware. Training experiments were conducted using open source model implementations from TensorFlow\cite{tf_models} and OpenSeq2Seq\cite{kuchaiev2018openseq2seq}. Our training framework internally updates the training graph by inserting quantization OPs, in the forward, backward, weight update paths for all convolution and GEMM kernels as described in Section.\ref{method}. 

Using the proposed training method, we have successfully trained Resnet-18, Resnet-34 and Resnet-50\cite{resnet_he} on Imagenet-1K\cite{imagenet_cvpr09} data set. We have used the same set of hyper parameters (except for loss scaling) and converged the network in the same number of iterations as the baseline FP32 training. For these convolution networks, the first convolution and the last fully-connected (FC) layers are maintained at a higher precision (16-bit) to maintain the model accuracy. 
For all convolution networks, in addition to using FP8 data format for weights, activations, error and weight gradients, we have also reduced the precision of the master copy of weights to FP16. Using techniques described in Section.\ref{noise_general}, we also manged to achieve slightly better top-1 accuracy compared to the baseline. Table.\ref{tab:top1_resnet} summarizes the validation accuracy achieved by convolution networks on imagenet-1K\cite{imagenet_cvpr09} dataset. 

\begin{table}[h] 
  \caption{Top-1 validation accuracy for convolution networks on Imagenet-1K\cite{imagenet_cvpr09} data set.}
  \label{tab:top1_resnet}
  \centering
  \begin{tabular}{llllll}
    \toprule
    Model     & Dataset     & Batch-size  & Epochs   & FP32 (top-1 \%)  & FP8 (top-1 \%)  \\
    \midrule
    \centering
    Resnet-18 & imagenet-1K & 256      & 100   & \num{69.23}  & \num{69.71}  \\
    Resnet-34 & imagenet-1K & 256      & 100   & \num{72.96}  & \num{72.95}  \\
    Resnet-50 & imagenet-1K & 256      & 100   & \num{75.47}  & \num{75.70}  \\
    \bottomrule
  \end{tabular}
\end{table}

\begin{figure}[!htb]
\centering
\begin{subfigure}{.495\textwidth}
  \centering
  \setlength{\fboxsep}{2pt}\fbox{\includegraphics[width=0.97\linewidth,keepaspectratio]{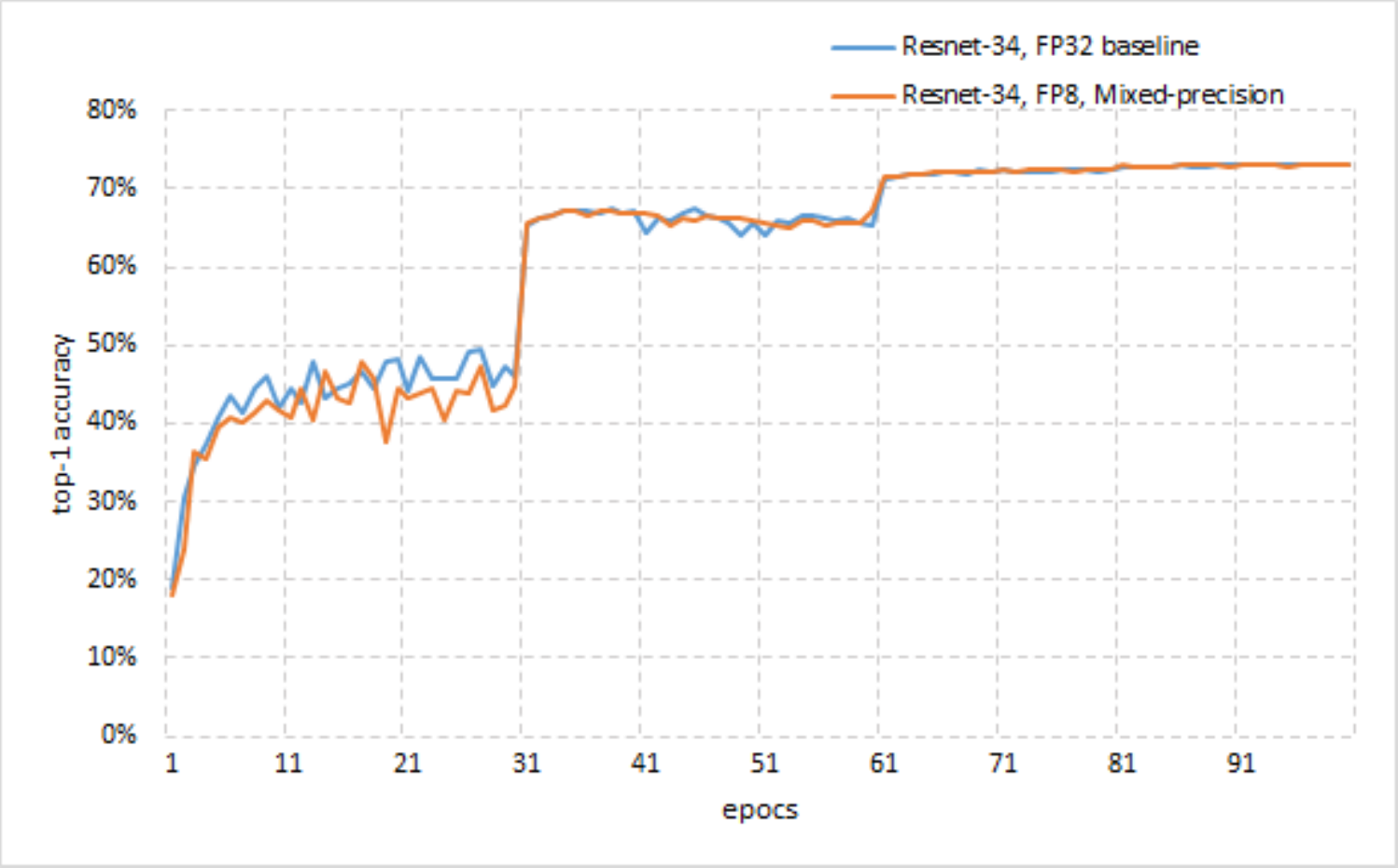}}
 \caption{}
  \label{fig:resnet34_top1}
\end{subfigure}
\begin{subfigure}{.495\textwidth}
  \centering
  \setlength{\fboxsep}{2pt}\fbox{\includegraphics[width=0.97\linewidth,keepaspectratio]{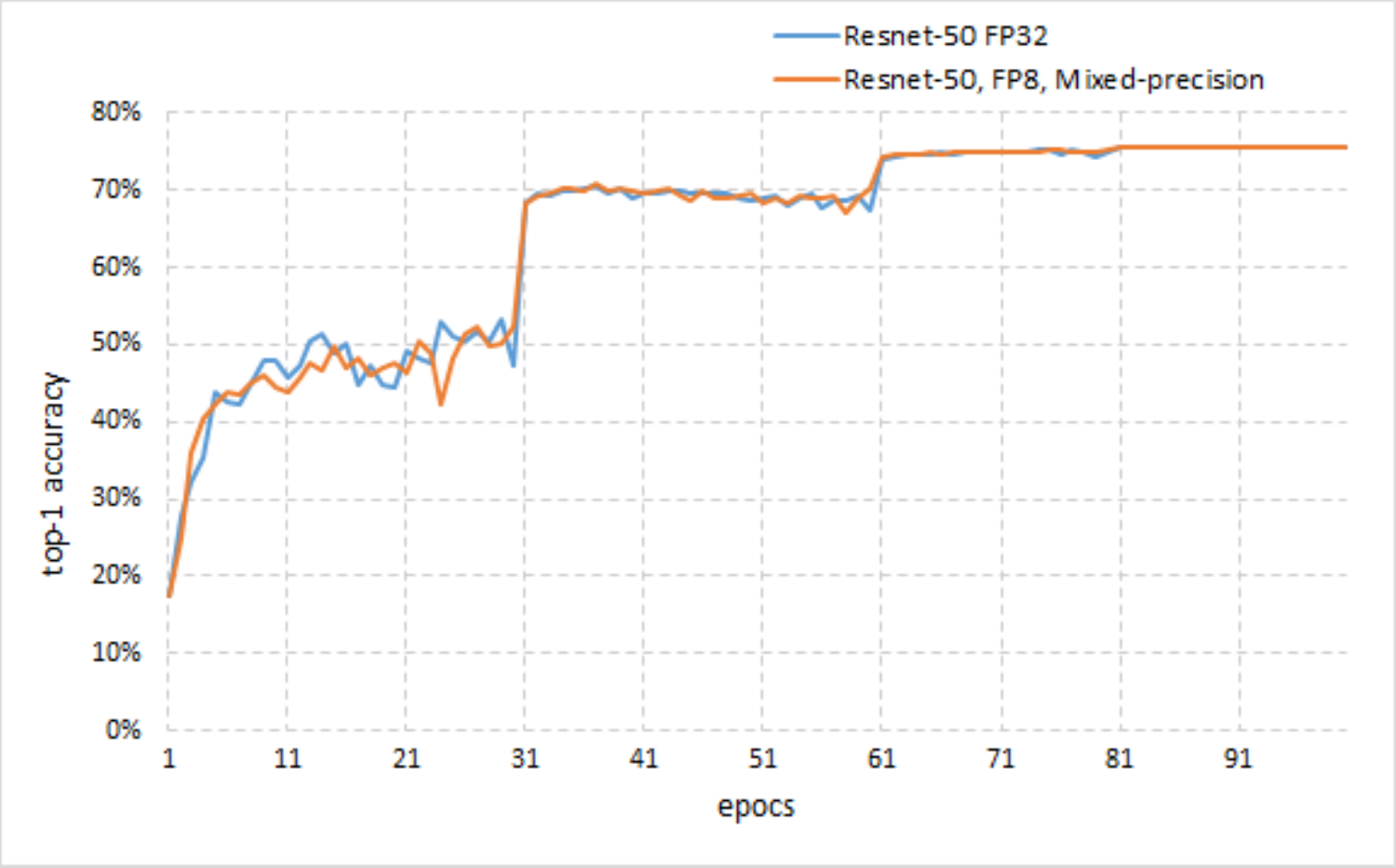}}
  \caption{}
  \label{fig:resnet50_top1}
\end{subfigure}
\caption{Convergence plots showing Top-1 validation accuracy for. (a) Resnet-34\cite{resnet_he} (b)  Resnet-50\cite{resnet_he} on imagenet-1K\cite{imagenet_cvpr09} dataset.}
\label{fig:resnet_plots}
\end{figure}

Figure.\ref{fig:resnet_plots} shows the convergence plots for Resnet-34 and Resnet-50 comparing top-1 accuracy of FP8 training with the baseline FP32 training. It can be seen that the validation accuracy of FP8 training closely follow the baseline numbers indicating the robustness of the training method.


\begin{table}[h] 
  \caption{Comparison of our method with the only other FP8 training method on Imagenet-1K\cite{imagenet_cvpr09} data set. W, A, E, G, MasterWts represent the precision setting for weights, activations, error, weight gradients and mater copy of weights respectively.}  
  \label{tab:competition}
  \centering
  \begin{tabular}{llllll}
    \toprule
    Method, Format     & W,A,E,G & MasterWts & Resnet-18 & Resnet-50 \\
                       &         &          & (top-1 error \%) & (top-1 error \%)  \\
    \midrule
    \centering
    Wang et al.\cite{NIPS2018_fp8}, FP8 & 8,8,8,8 & 16 & \num{ 33.05}   & \num{28.28}  \\
    Ours, FP8 & 8,8,8,8  & 16 & \num{30.29} & \num{24.30} \\
    \bottomrule
  \end{tabular}
\end{table}

\begin{figure}[!htb]
\centering
\begin{subfigure}{.495\textwidth}
  \centering
  \setlength{\fboxsep}{2pt}\fbox{\includegraphics[width=0.97\linewidth,keepaspectratio]{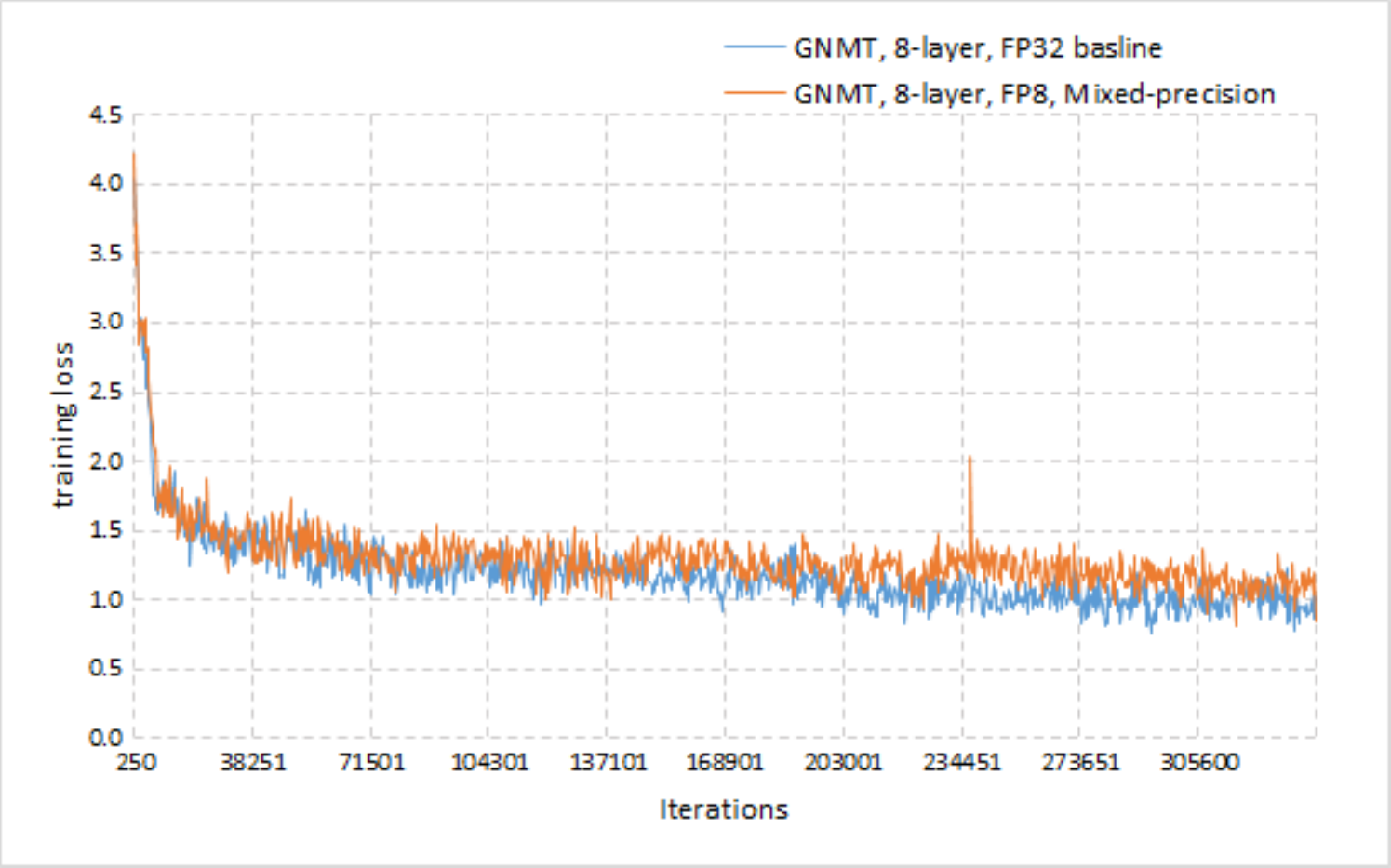}}
 \caption{}
  \label{fig:gnmt_trainloss}
\end{subfigure}
\begin{subfigure}{.495\textwidth}
  \centering
  \setlength{\fboxsep}{2pt}\fbox{\includegraphics[width=0.97\linewidth,keepaspectratio]{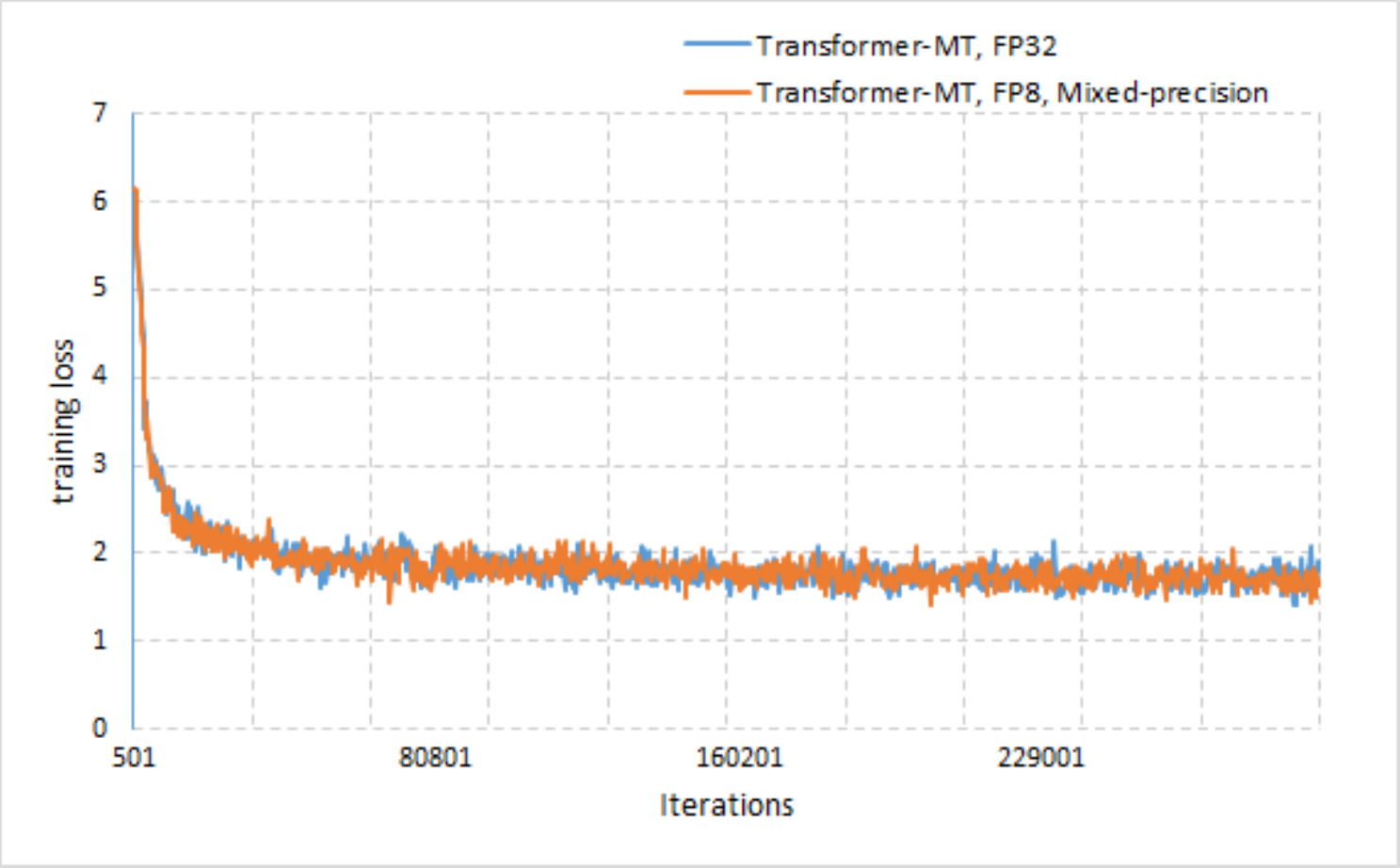}}
  \caption{}
  \label{fig:transformer_trainloss}
\end{subfigure}
\vspace{-5px}
\caption{Convergence plots showing training loss for (a) 8-layer GNMT\cite{google_gnmt} and, (b) 6-layer Transformer\cite{attention_transformer} trained on WMT16 English->German dataset.}
\label{fig:gnmt_plots}
\end{figure}
In addition to convolution networks, we have also trained two state of the art machine translation workloads (GNMT\cite{google_gnmt} and Transformer\cite{attention_transformer}) and demonstrated BLEU scores matching single precision baselines. We trained an 8-layer GNMT\cite{google_gnmt} encoder/decoder LSTM model with 1024 recurrent units and 1024 attention units. 
We trained this network using FP8 numeric format for all GEMM operations, while the activation functions such as tanh and sigmoid use FP16 data type. We used the loss scaling schedule described in Section.\ref{loss_scaling}. 

We also trained a 6-layer Transformer\cite{attention_transformer} translation network with with roughly 200M parameters. 
For the Transformer network, our internal baseline score is lower than the current reported highest score. Both GNMT\cite{google_gnmt} and Transformer\cite{attention_transformer} models were trained on large scale, WMT2016 English$\xrightarrow{ }$German dataset consisting of 4.5 million sentence pairs. We trained these networks using ADAM optimizer with same hyper parameters used by the FP32 baseline. On both these models, our FP8 mixed precision training achieved BLEU score comparable to the FP32 baseline. The results are summarized in Table.\ref{tab:bleu_score}.
 
\begin{table}[h] 
  \caption{sacreBLEU\cite{sacre_bleu} score measured on WMT 2014 English$\xrightarrow{}$German dataset}
  \label{tab:bleu_score}
  \centering
  \begin{tabular}{llll}
    \toprule
    Model   & Dataset/ Task & FP32 baseline & FP8 Mixed Precision  \\
    \midrule
    \centering
    GNMT        & WMT 2016 English$\xrightarrow{}$German & \num{24.3}  & \num{24.6}  \\
    Transformer & WMT 2016 English$\xrightarrow{}$German  & \num{23.6}  & \num{23}  \\
    \bottomrule
  \end{tabular}
\end{table}

\section{Conclusion}\label{conclusion}
We demonstrate state-of-the-art accuracy across multiple data sets (imagenet-1K, WMT16) and a broader set of workloads (Resnet-18/34/50, GNMT, Transformer) than previously reported. We propose easy to implement and scalable solution for building FP8 compute primitives, eliminating the need for stochastic rounding hardware in the critical compute path, as proposed in \cite{NIPS2018_fp8}, thereby reducing the cost and complexity of the MAC unit. We explore issues around gradient underflow and quantization noise that arise as a result of using the proposed 8-bit numeric format for large scale neural network training. We propose solutions to deal with these problems in the form of enhanced loss scaling and stochastic rounding.


\begin{thebibliography}{10}

\bibitem{tf_models}
{Models and examples built with TensorFlow}.
\newblock \url{https://github.com/tensorflow/models}.

\bibitem{tf_framework}
{Tensorflow framework for reduced precision training}.
\newblock \url{https://github.com/nkmellem/tensorflow}.

\bibitem{ai_and_compute}
Dario Amodei and Danny Hernandez.
\newblock {AI and Compute}.
\newblock \url{https://openai.com/blog/ai-and-compute/}.

\bibitem{halfwave_gaussian}
Zhaowei Cai, Xiaodong He, Jian Sun, and Nuno Vasconcelos.
\newblock Deep learning with low precision by half-wave gaussian quantization.
\newblock In {\em Proceedings of the IEEE Conference on Computer Vision and
  Pattern Recognition}, pages 5918--5926, 2017.

\bibitem{int16_intel}
Dipankar Das, Naveen Mellempudi, Dheevatsa Mudigere, Dhiraj Kalamkar, Sasikanth
  Avancha, Kunal Banerjee, Srinivas Sridharan, Karthik Vaidyanathan, Bharat
  Kaul, Evangelos Georganas, et~al.
\newblock Mixed precision training of convolutional neural networks using
  integer operations.
\newblock {\em arXiv preprint arXiv:1802.00930}, 2018.

\bibitem{8b_halp}
Christopher De~Sa, Megan Leszczynski, Jian Zhang, Alana Marzoev, Christopher~R
  Aberger, Kunle Olukotun, and Christopher R{\'e}.
\newblock High-accuracy low-precision training.
\newblock {\em arXiv preprint arXiv:1803.03383}, 2018.

\bibitem{imagenet_cvpr09}
J.~Deng, W.~Dong, R.~Socher, L.-J. Li, K.~Li, and L.~Fei-Fei.
\newblock {ImageNet: A Large-Scale Hierarchical Image Database}.
\newblock In {\em CVPR09}, 2009.

\bibitem{Glorot}
Xavier Glorot and Yoshua Bengio.
\newblock Understanding the difficulty of training deep feedforward neural
  networks.
\newblock In {\em In Proceedings of the International Conference on Artificial
  Intelligence and Statistics (AISTATS’10). Society for Artificial
  Intelligence and Statistics}, 2010.

\bibitem{speech_hinton}
Alex Graves, Abdel-rahman Mohamed, and Geoffrey Hinton.
\newblock Speech recognition with deep recurrent neural networks.
\newblock In {\em 2013 IEEE international conference on acoustics, speech and
  signal processing}, pages 6645--6649. IEEE, 2013.

\bibitem{gupta_16bit}
Suyog Gupta, Ankur Agrawal, Kailash Gopalakrishnan, and Pritish Narayanan.
\newblock Deep learning with limited numerical precision.
\newblock In {\em International Conference on Machine Learning}, pages
  1737--1746, 2015.

\bibitem{deep_speech}
Awni Hannun, Carl Case, Jared Casper, Bryan Catanzaro, Greg Diamos, Erich
  Elsen, Ryan Prenger, Sanjeev Satheesh, Shubho Sengupta, Adam Coates, et~al.
\newblock Deep speech: Scaling up end-to-end speech recognition.
\newblock {\em arXiv preprint arXiv:1412.5567}, 2014.

\bibitem{resnet_he}
Kaiming He, Xiangyu Zhang, Shaoqing Ren, and Jian Sun.
\newblock Deep residual learning for image recognition.
\newblock In {\em Proceedings of the IEEE conference on computer vision and
  pattern recognition}, pages 770--778, 2016.

\bibitem{hubara_bnn}
Itay Hubara, Matthieu Courbariaux, Daniel Soudry, Ran El-Yaniv, and Yoshua
  Bengio.
\newblock Quantized neural networks: Training neural networks with low
  precision weights and activations.
\newblock {\em The Journal of Machine Learning Research}, 18(1):6869--6898,
  2017.

\bibitem{pmr}
Atef Ibrahim and Fayez Gebali.
\newblock Low power semi-systolic architectures for polynomial-basis
  multiplication over gf (2 m) using progressive multiplier reduction.
\newblock {\em Journal of Signal Processing Systems}, 82(3):331--343, 2016.

\bibitem{keskar2016large}
Nitish~Shirish Keskar, Dheevatsa Mudigere, Jorge Nocedal, Mikhail Smelyanskiy,
  and Ping Tak~Peter Tang.
\newblock On large-batch training for deep learning: Generalization gap and
  sharp minima.
\newblock {\em arXiv preprint arXiv:1609.04836}, 2016.

\bibitem{flexpoint}
Urs K{\"o}ster, Tristan Webb, Xin Wang, Marcel Nassar, Arjun~K Bansal, William
  Constable, Oguz Elibol, Scott Gray, Stewart Hall, Luke Hornof, et~al.
\newblock Flexpoint: An adaptive numerical format for efficient training of
  deep neural networks.
\newblock In {\em Advances in neural information processing systems}, pages
  1742--1752, 2017.

\bibitem{koster2017flexpoint}
Urs K{\"o}ster, Tristan Webb, Xin Wang, Marcel Nassar, Arjun~K Bansal, William
  Constable, Oguz Elibol, Scott Gray, Stewart Hall, Luke Hornof, et~al.
\newblock Flexpoint: An adaptive numerical format for efficient training of
  deep neural networks.
\newblock In {\em Advances in neural information processing systems}, pages
  1742--1752, 2017.

\bibitem{alexnet}
Alex Krizhevsky, Ilya Sutskever, and Geoffrey~E Hinton.
\newblock Imagenet classification with deep convolutional neural networks.
\newblock In {\em Advances in neural information processing systems}, pages
  1097--1105, 2012.

\bibitem{kuchaiev2018openseq2seq}
Oleksii Kuchaiev, Boris Ginsburg, Igor Gitman, Vitaly Lavrukhin, Jason Li,
  Huyen Nguyen, Carl Case, and Paulius Micikevicius.
\newblock Mixed-precision training for nlp and speech recognition with
  openseq2seq.
\newblock {\em Computing Research Repository (CoRR), abs/1805.10387 v2}, 2018.

\bibitem{nvidia_tensorcores}
Stefano Markidis, Steven~Wei Der~Chien, Erwin Laure, Ivy~Bo Peng, and Jeffrey~S
  Vetter.
\newblock Nvidia tensor core programmability, performance \& precision.
\newblock In {\em 2018 IEEE International Parallel and Distributed Processing
  Symposium Workshops (IPDPSW)}, pages 522--531. IEEE, 2018.

\bibitem{fp16_mixed_nv}
Paulius Micikevicius, Sharan Narang, Jonah Alben, Gregory Diamos, Erich Elsen,
  David Garcia, Boris Ginsburg, Michael Houston, Oleksii Kuchaiev, Ganesh
  Venkatesh, et~al.
\newblock Mixed precision training.
\newblock {\em arXiv preprint arXiv:1710.03740}, 2017.

\bibitem{sacre_bleu}
Matt Post.
\newblock A call for clarity in reporting bleu scores.
\newblock {\em arXiv preprint arXiv:1804.08771}, 2018.

\bibitem{attention_transformer}
Ashish Vaswani, Noam Shazeer, Niki Parmar, Jakob Uszkoreit, Llion Jones,
  Aidan~N Gomez, {\L}ukasz Kaiser, and Illia Polosukhin.
\newblock Attention is all you need.
\newblock In {\em Advances in neural information processing systems}, pages
  5998--6008, 2017.

\bibitem{NIPS2018_fp8}
Naigang Wang, Jungwook Choi, Daniel Brand, Chia-Yu Chen, and Kailash
  Gopalakrishnan.
\newblock Training deep neural networks with 8-bit floating point numbers.
\newblock In S.~Bengio, H.~Wallach, H.~Larochelle, K.~Grauman, N.~Cesa-Bianchi,
  and R.~Garnett, editors, {\em Advances in Neural Information Processing
  Systems 31}, pages 7675--7684. Curran Associates, Inc., 2018.

\bibitem{iclr17_int8}
Shuang Wu, Guoqi Li, Feng Chen, and Luping Shi.
\newblock Training and inference with integers in deep neural networks.
\newblock In {\em International Conference on Learning Representations}, 2018.

\bibitem{google_gnmt}
Yonghui Wu, Mike Schuster, Zhifeng Chen, Quoc~V Le, Mohammad Norouzi, Wolfgang
  Macherey, Maxim Krikun, Yuan Cao, Qin Gao, Klaus Macherey, et~al.
\newblock Google's neural machine translation system: Bridging the gap between
  human and machine translation.
\newblock {\em arXiv preprint arXiv:1609.08144}, 2016.

\bibitem{zhou2016dorefa}
Shuchang Zhou, Yuxin Wu, Zekun Ni, Xinyu Zhou, He~Wen, and Yuheng Zou.
\newblock Dorefa-net: Training low bitwidth convolutional neural networks with
  low bitwidth gradients.
\newblock {\em arXiv preprint arXiv:1606.06160}, 2016.

\bibitem{neural_arch_search}
Barret Zoph and Quoc~V. Le.
\newblock Neural architecture search with reinforcement learning.
\newblock {\em CoRR}, abs/1611.01578, 2016.

\end{thebibliography}
\end{document}